\documentclass[pmlr]{jmlr}% new name PMLR (Proceedings of Machine Learning)

\usepackage{longtable}% for long tables

\usepackage{booktabs}
\usepackage[load-configurations=version-1]{siunitx} % newer version
\theorembodyfont{\upshape}
\theoremheaderfont{\scshape}
\theorempostheader{:}
\theoremsep{\newline}

\jmlrvolume{1}
\jmlryear{2017}
\jmlrworkshop{ICML 2017 AutoML Workshop}

\title[Dealing with Integer-valued Variables in Bayesian Optimization]
{Dealing with Integer-valued Variables in Bayesian Optimization with Gaussian Processes}

  \author{\Name{Eduardo C. Garrido-Merch\'an} \Email{eduardo.garrido@uam.es} \\
	\Name{Daniel Hern\'andez-Lobato} \Email{daniel.hernandez@uam.es}\\
   \addr Universidad Aut\'onoma de Madrid, Francisco Tom\'as y Valiente 11, 28049, Madrid, Spain}

\begin{document}

\maketitle

\begin{abstract}
Bayesian optimization (BO) methods are useful for optimizing functions
that are expensive to evaluate, lack an analytical expression and whose evaluations can 
be contaminated by noise. These methods rely on a probabilistic model of the objective
function, typically a Gaussian process (GP), upon which an acquisition function is built.
This function guides the optimization process and measures the expected utility of performing 
an evaluation of the objective at a new point. GPs assume continous input variables. When this 
is not the case, such as when some of the input variables take integer values, one has to introduce 
extra approximations. A common approach is to round the suggested variable value to the closest integer 
before doing the evaluation of the objective. We show that this can lead to problems in the optimization 
process and describe a more principled approach to account for input variables that are integer-valued.  
We illustrate in both synthetic and a real experiments the utility of our approach, which significantly 
improves the results of standard BO methods on problems involving integer-valued variables.
\end{abstract}


\section{Background on Bayesian Optimization} \label{sec:background_bo}

Bayesian optimization (BO) methods \citep{shahriari2016} address the problem of optimizing a real-valued 
function $f(\mathbf{x})$ over some bounded domain $\mathcal{X}$. 
The objective function is assumed to lack an analytical expression (which prevents any gradient computation),
to be very expensive to evaluate, and the evaluations are assumed to be noisy (\emph{i.e.}, rather than
observing $f(\mathbf{x})$ we observe $y = f(\mathbf{x}) + \epsilon$, with $\epsilon$ some additive noise). 
At each iteration $t=1,2,3,\ldots$ of the optimization process, BO methods fit a probabilistic model, 
typically a Gaussian process (GP) \citep{rasmussen2005book}, to the observations of the objective 
function $\{y_i\}_{i=1}^{t-1}$ collected so far.
The uncertainty about the potential values of the objective function provided by the GP is then used to 
generate an acquisition function $\alpha(\cdot)$, whose value at each input location indicates the expected utility 
of evaluating $f(\cdot)$ there. The next point $\mathbf{x}_t$ at which to evaluate $f(\cdot)$ is the 
one that maximizes $\alpha(\cdot)$.  After collecting this observation, the process 
is repeated. When enough data has been collected, the GP predictive mean value 
for $f(\cdot)$ can be optimized to find the solution of the problem.

The key for BO success is that evaluating the acquisition function $\alpha(\cdot)$ is very cheap compared
to the evaluation of $f(\cdot)$. This is so because the acquisition function only depends on 
the GP predictive distribution for $f(\cdot)$ at a candidate point $\mathbf{x}$. Thus, $\alpha(\cdot)$ can be maximized 
with very little cost. BO methods hence spend a small amount 
of time thinking very carefully where to evaluate next the objective function with the aim of finding its optimum with 
the smallest number of evaluations. This is a very useful strategy when the objective function is very expensive to 
evaluate and it can save a lot of computational time.  

Let the observed data until step $t-1$ of the algorithm be $\mathcal{D}_i=\{(\mathbf{x}_i,y_i)\}_{i=1}^{t-1}$. 
The GP predictive distribution for $f(\cdot)$
is given by a Gaussian distribution characterized by a mean $\mu(\mathbf{x})$ and a variance $\sigma^2(\mathbf{x})$. 
These values are:
\begin{align}
\mu(\mathbf{x}) & = \mathbf{k}_{*}^{T} (\mathbf{K}+\sigma_{0}^{2}\mathbf{I})^{-1}\mathbf{y}\,, 
& \sigma^2(\mathbf{x}) & = k(\mathbf{x},\mathbf{x}) - \mathbf{k}_{*}^T(\mathbf{K}+\sigma_0^2\mathbf{I})^{-1}\mathbf{k}_*\,,
\end{align}
where $\mathbf{y}=(y_1,\ldots,y_{t-1})^\text{T}$ is a vector with the objective values observed so far;
$\mathbf{k}_*$ is a vector with the prior covariances between $f(\mathbf{x})$ and each $y_i$;
$\sigma_0^2$ is the variance of the additive Gaussian noise;
$\mathbf{K}$ is a matrix with the prior covariances among each $f(\mathbf{x}_i)$, for $i=1,\ldots,t-1$;
and $k(\mathbf{x},\mathbf{x})$ is the prior variance at the candidate location $\mathbf{x}$.
These quantities are obtained from a covariance function $k(\cdot,\cdot)$ which is pre-specified 
and receives as an input $\mathbf{x}_i$ and $\mathbf{x}_j$
at which the covariance between $f(\mathbf{x}_i)$ and $f(\mathbf{x}_j)$ is evaluated.
A typical covariance function employed in BO is the Mat\'ern function \citep{Snoek2012}.

A popular acquisition function is expected improvement (EI) \citep{Jones98}. EI is given by the expected
value of the utility function $u(y) = \text{max}{(0, \nu - y)}$ under the GP predictive distribution for $y$,
where $\nu=\text{min}(\{y_i\}_{i=1}^{t-1})$ is the best value observed so far (assuming minimization). 
Thus, EI measures on average how much we will improve on the current best solution by performing an evaluation at each candidate point. 
The EI acquisition function is 
$\alpha(\mathbf{x}) = \sigma(\mathbf{x})(\gamma(\mathbf{x}) \Phi(\gamma(\mathbf{x}) + \phi(\gamma(\mathbf{x}))$,
where $\gamma(\mathbf{x}) = (\nu - \mu(\mathbf{x})) /\sigma(\mathbf{x})$ and $\Phi(\cdot)$ and $\phi(\cdot)$ are
respectively the c.d.f. and p.d.f. of a standard Gaussian.

\section{Dealing with Integer-valued Variables} \label{sec:dealing}

The framework described assumes continous input variables in $f(\cdot)$. This is so because in a GP
the variables introduced in the covariance function are assumed to be continous. A problem may arise 
when some of the inputs can only take values in a closed subset of a discrete set, such as the 
integers. If this is the case, the GP process will ignore that constraint and will place some probability 
mass on invalid potential values for $f(\cdot)$. These incorrect modeling assumptions about the objective
$f(\cdot)$ may have a negative impact on the optimization process. Furthermore, the optimization 
of $\alpha(\cdot)$ will typically provide candidate points at which to evaluate the objective 
$f(\cdot)$ that are invalid in the sense that integer-valued input variables will be assigned real values. 
In practice, some mechanism to transform real values into integer values must be implemented before the 
evaluation can take place. If this is not done with care, some problems may appear in the optimization process.

Optimization problems involving continuous and discrete variables appear in the task of optimizing 
the hyper-parameters of machine learning systems \citep{Snoek2012}. For example, in a deep neural network 
we may be interested in adjusting the learning rate, the number of layers and the number of neurons per layer,
which can only take discrete values. Similarly, in a ensemble of decision trees generated by the 
gradient boosting algorithm \citep{friedman2001greedy} we may try to adjust the learning rate and 
the maximum depth of the trees. This last hyper-parameter can only take discrete values. A last
example involves a nearest neighbor classifier. In this case we may be interested in finding the optimal number 
of neighbors and the optimal scaling factor per dimension to be used in the computation of the distance. The number 
of neighbors can only take discrete values.

\begin{figure}[htb]
\begin{tabular}{l@{\hspace{1mm}}cc}
	\rotatebox{90}{\hspace{.7cm}{\bf \scriptsize Naive}} &  
        \includegraphics[width=0.475\linewidth]{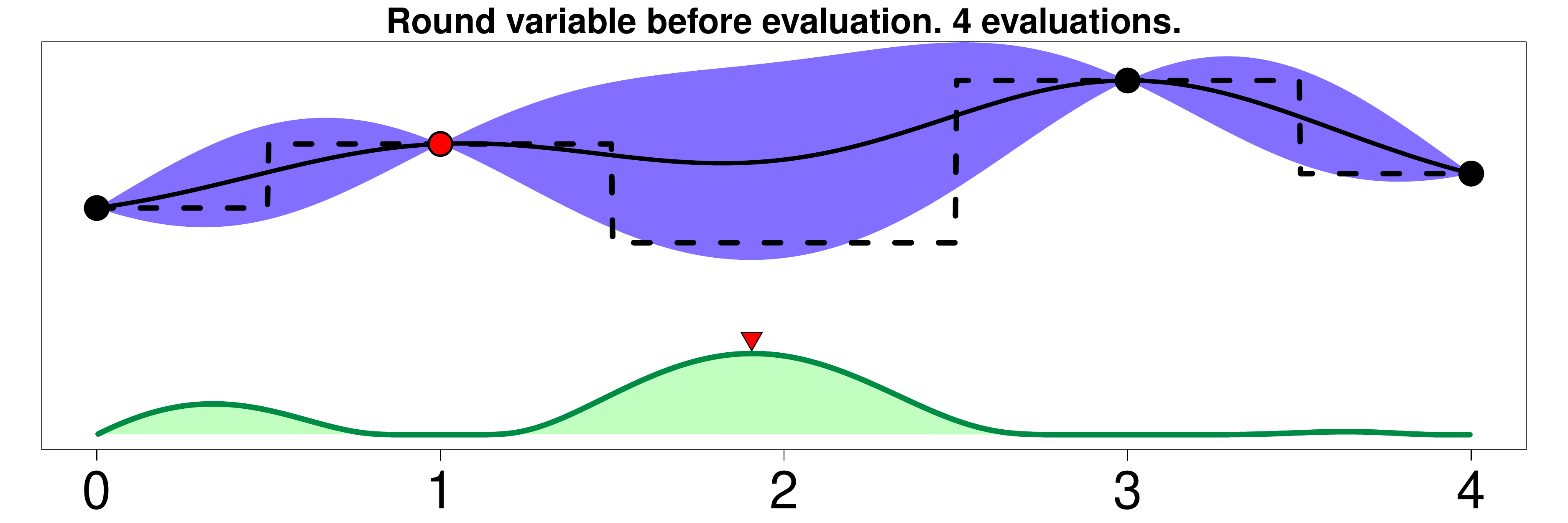} &
        \includegraphics[width=0.475\linewidth]{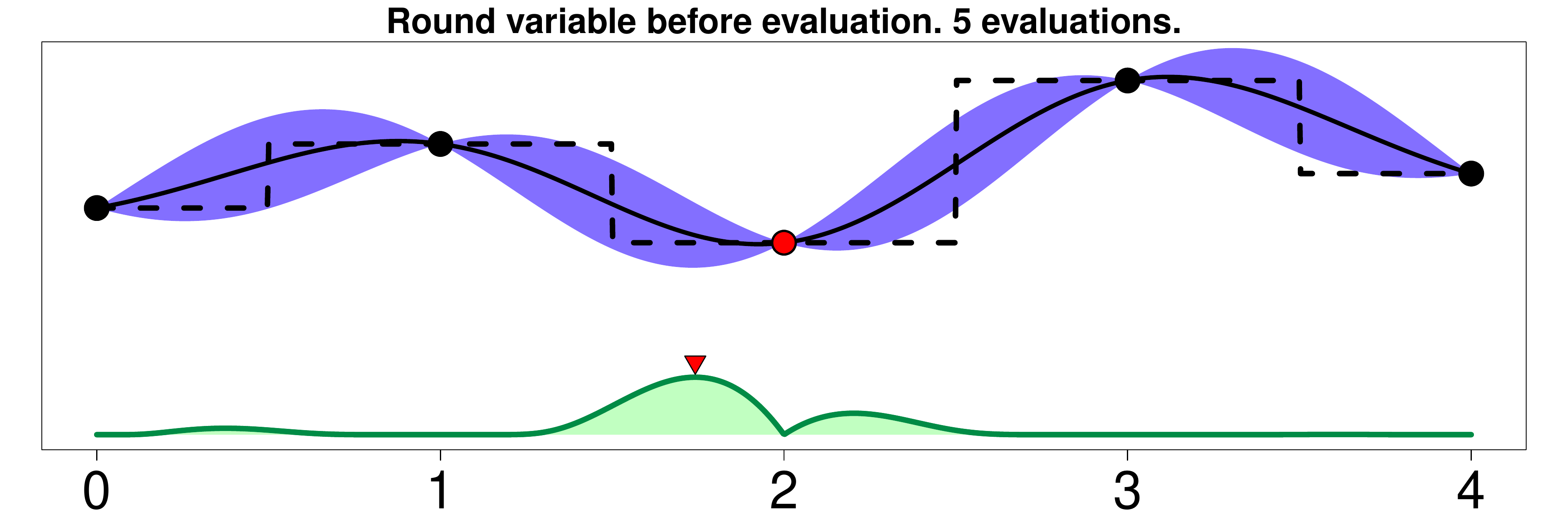} \\
	\rotatebox{90}{\hspace{.7cm}{\bf \scriptsize Basic}} &  
        \includegraphics[width=0.475\linewidth]{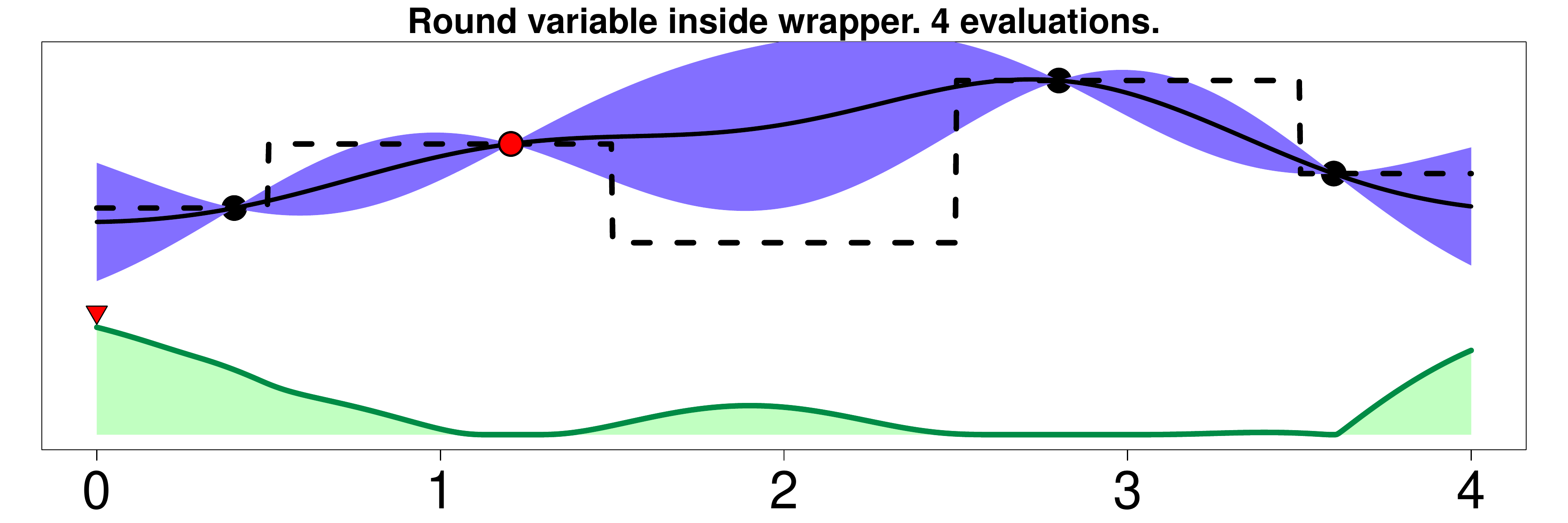} &
        \includegraphics[width=0.475\linewidth]{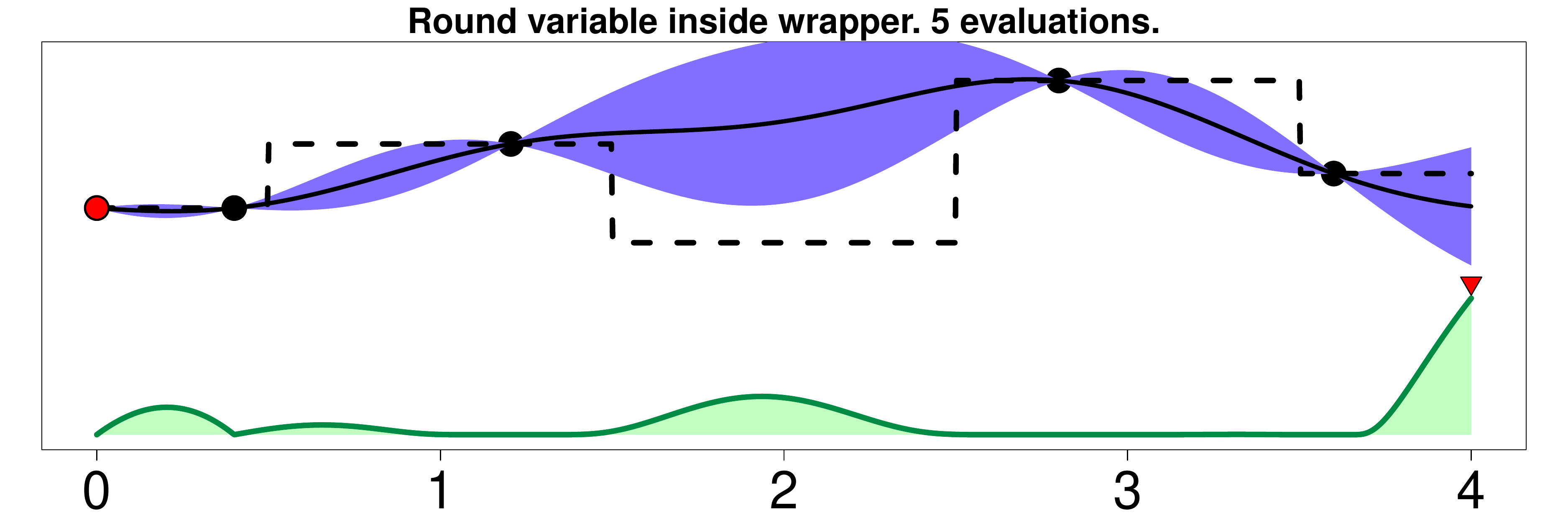} \\
	\rotatebox{90}{\hspace{.6cm}{\bf \scriptsize Proposed}} &  
        \includegraphics[width=0.475\linewidth]{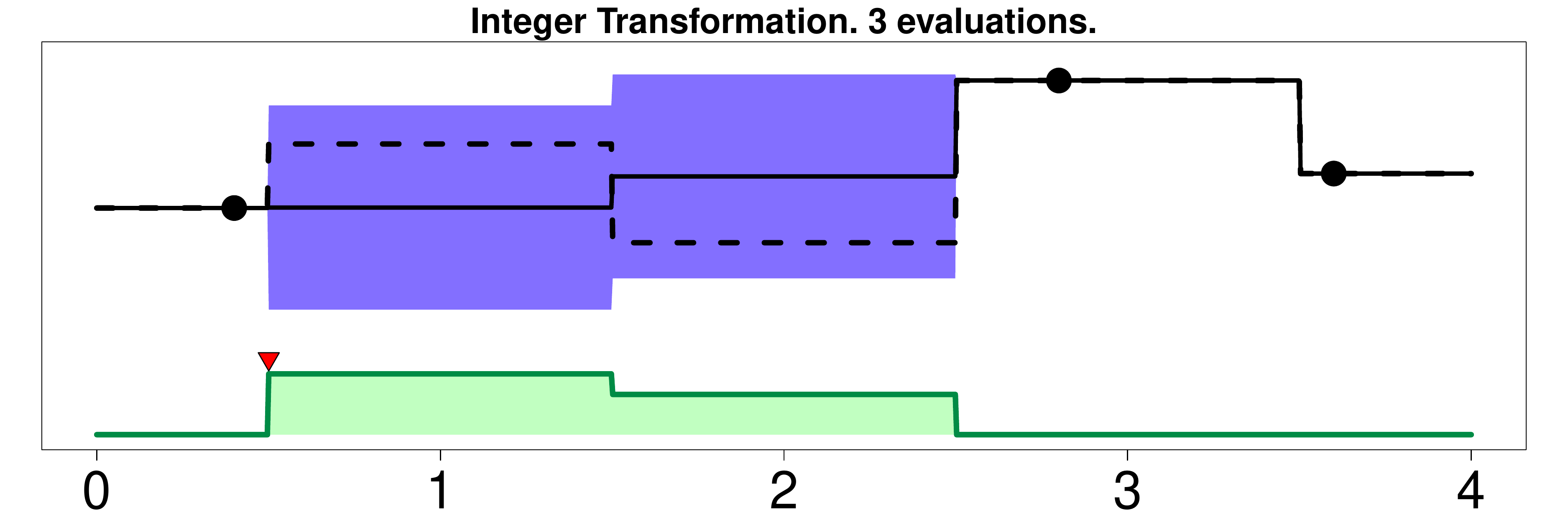} &
        \includegraphics[width=0.475\linewidth]{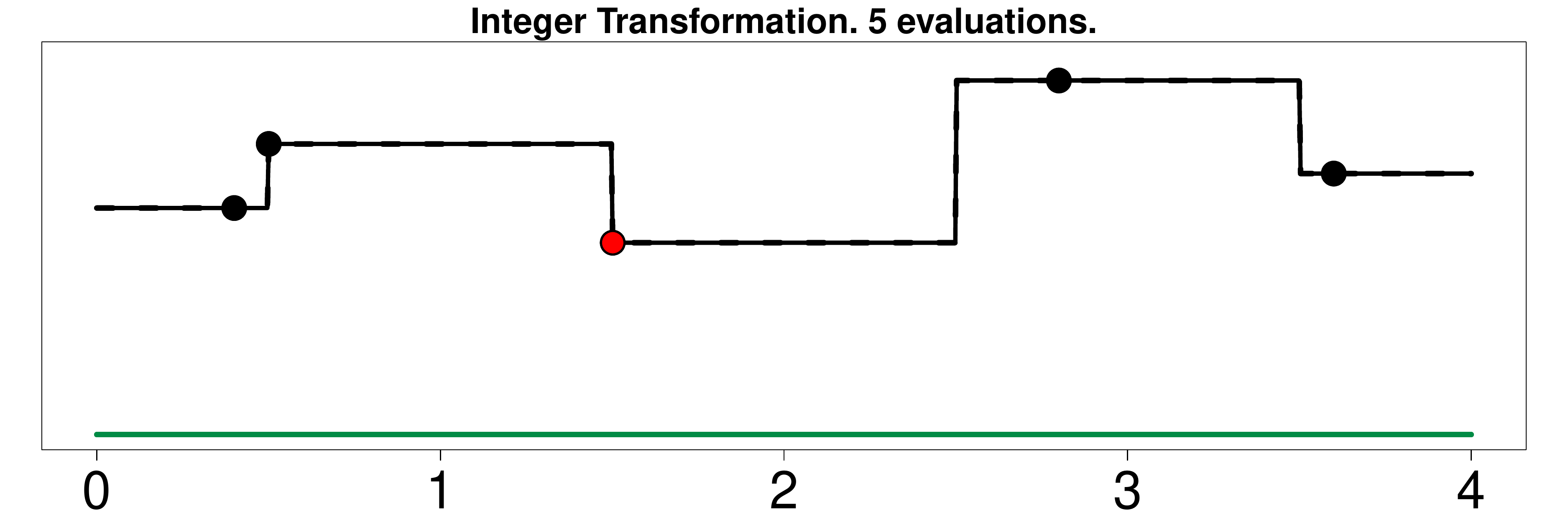} \\
\end{tabular}
\vspace{-.5cm}
\caption{{\small Different methods for dealing with integer-valued variables. 
At the top of each image we show a GP fit to the data (posterior mean and 1-std confidence interval, in purple) 
that models a 1-dimensional objective taking values in the set $\{0,1,2,3,4\}$ (dashed line). 
To display the objective we have rounded the real values at which to do the evaluation to the closest 
integer. Below the GP fit it is shown the acquisition function whose maximum is the recommendation for the next new 
evaluation. Each column show similar figures before and after evaluating a new point, respectively. 
The proposed approach leads to no uncertainty about the objective after two evaluations.  Best seen in color.}}
\label{fig:methods}
\vspace{-.5cm}
\end{figure}

A naive approach to consider that the objective can only be evaluated at integer values in some of the inputs 
is to (i) optimize $\alpha(\cdot)$ assuming all variables take values in the real line, and (ii) replace all 
the values for the integer-valued variables by the closest integer. This is the approach followed by 
the popular software for BO Spearmint ({\small \url{https://github.com/HIPS/Spearmint}}). However, as shown in 
the first row of Figure \ref{fig:methods} this can lead to a mismatch between the points in which the 
acquisition takes high values, and where the actual evaluation is performed. Furthermore, it can produce situations in 
which the BO method always evaluates the objective at a point where it has already been evaluated (because the next
and following evaluations will not change at all the acquisition function). For this reason, we discourage the use of this 
approach.

The previous problem can be solved by doing the rounding to the closest integer value inside the 
wrapper that evaluates the objective. This basic approach is shown in the second row of Figure \ref{fig:methods}.
In this case the points at which the acquisition takes high values and the points at which the
objective is evaluated coincide. Thus, the BO method will tend to evaluate at different locations,
as expected. The problem is, however, that the actual objective is constant in the intervals that
are rounded to the same integer value. This constant behavior is ignored by the GP, which can be sub-optimal.

\section{Proposed Approach} \label{sec:proposed}

We propose a method to alleviate the problems of the basic approach of Section \ref{sec:dealing}.
For this, we consider that the objective should be constant in the intervals that are rounded to 
the same integer. This property can be easily introduced in the GP by modifying
$k(\cdot,\cdot)$. Covariance functions are often stationary and only depend on the distance between the 
input points. If the distance between two points its zero, the values of the function at both 
points will be the same (the correlation is equal to one). Based on this fact, we suggest to transform the input 
points to $k(\cdot,\cdot)$, obtaining an alternative covariance function $k'(\cdot,\cdot)$:
\begin{align}
k'(\mathbf{x}_i,\mathbf{x}_j) &= k(T(\mathbf{x}_i),T(\mathbf{x}_j)) \,,
\label{eq:covariance}
\end{align}
where $T(\mathbf{x})$ is a transformation in which all integer-valued variables of $f(\cdot)$ in $\mathbf{x}$ are rounded to 
the closest integer. The beneficial properties of $k'(\cdot,\cdot)$ when used for BO are illustrated in the 
third row of Figure \ref{fig:methods}. We can see that the GP model correctly identifies that the objective 
function is constant inside intervals of real values that are rounded to the same integer. The uncertainty is 
also the same in those intervals, and this is reflected in the acquisition function. Furthermore, after performing a 
single measurement in each interval, the uncertainty about $f(\cdot)$ goes to zero. This better modeling
of the objective is expected to be reflected in a better performance of the optimization process.

\begin{figure}[htb]
\begin{center}
\begin{tabular}{lcc}
	\rotatebox{90}{\hspace{1.75cm}{\bf \scriptsize Proposed Approach}} &  
        \includegraphics[width=0.375\linewidth]{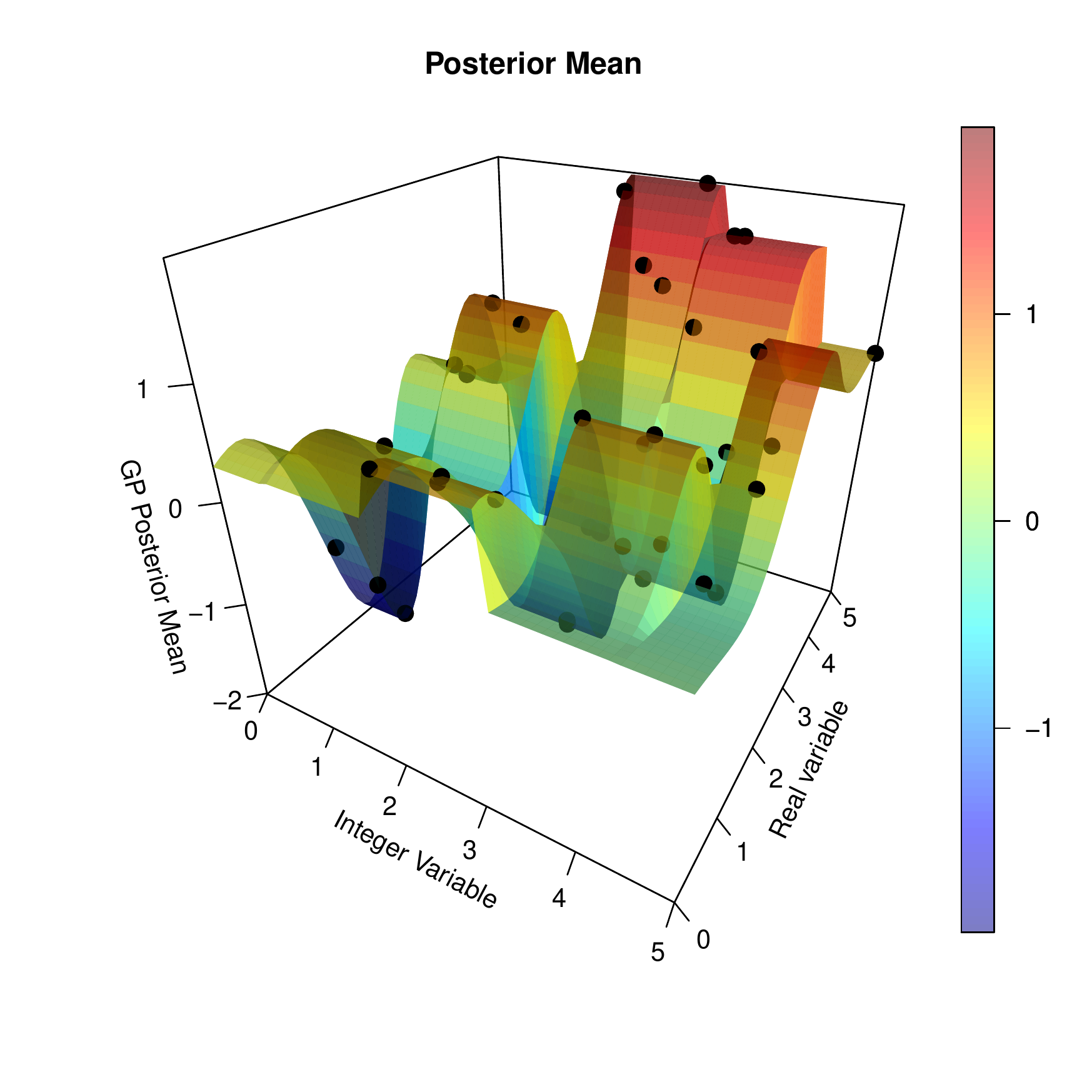} &
        \includegraphics[width=0.375\linewidth]{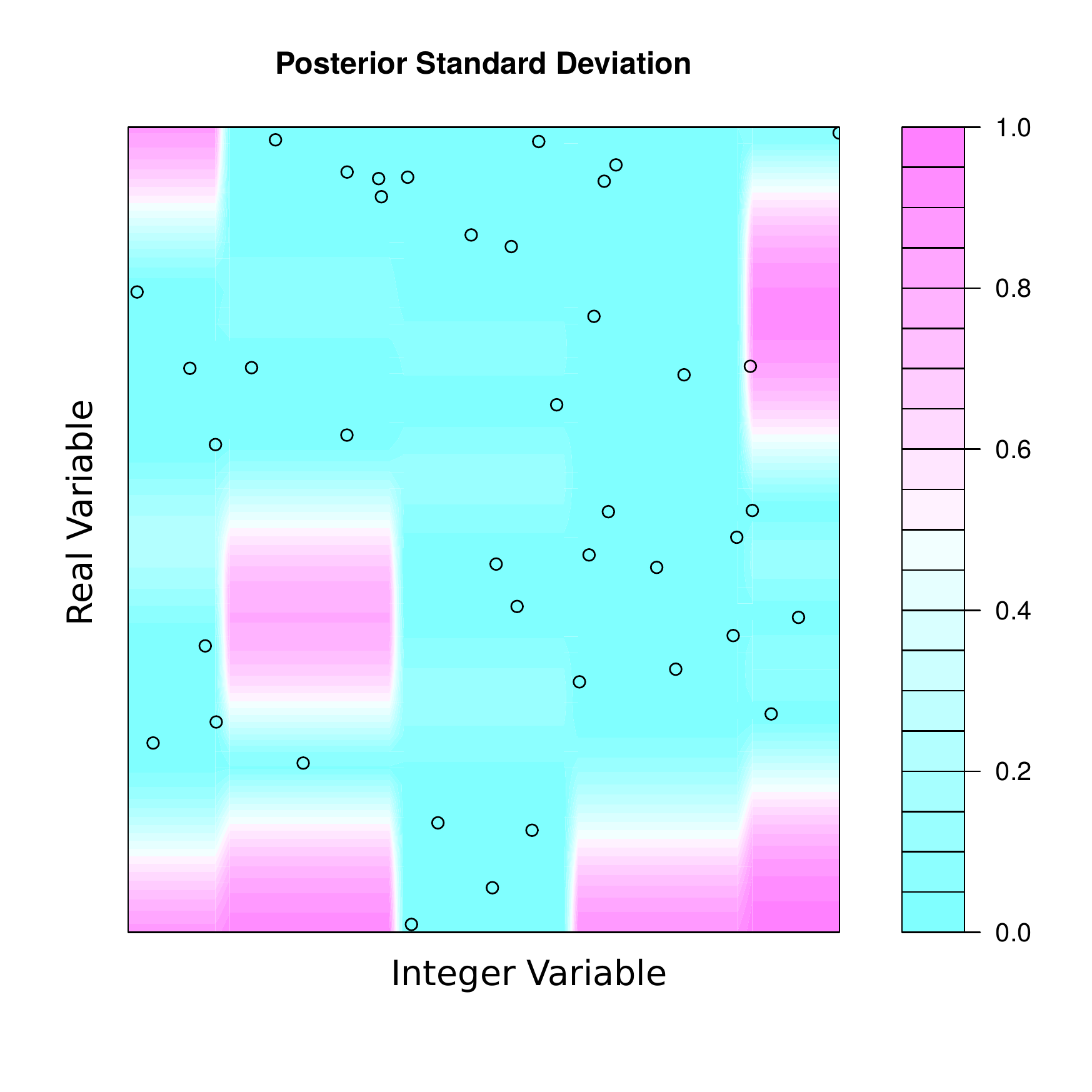} \\
	\rotatebox{90}{\hspace{2.25cm}{\bf \scriptsize Standard GP}} &  
        \includegraphics[width=0.375\linewidth]{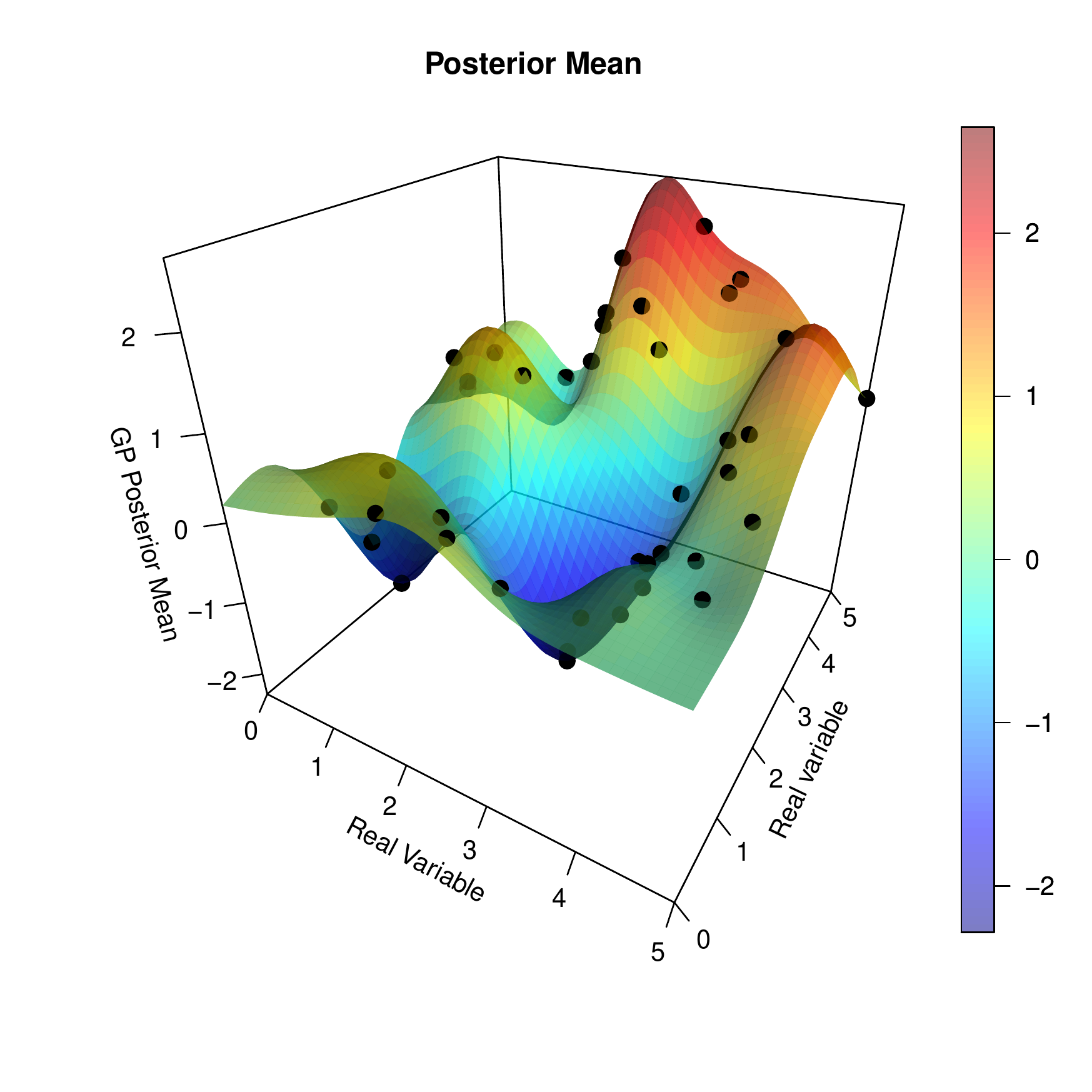} &
        \includegraphics[width=0.375\linewidth]{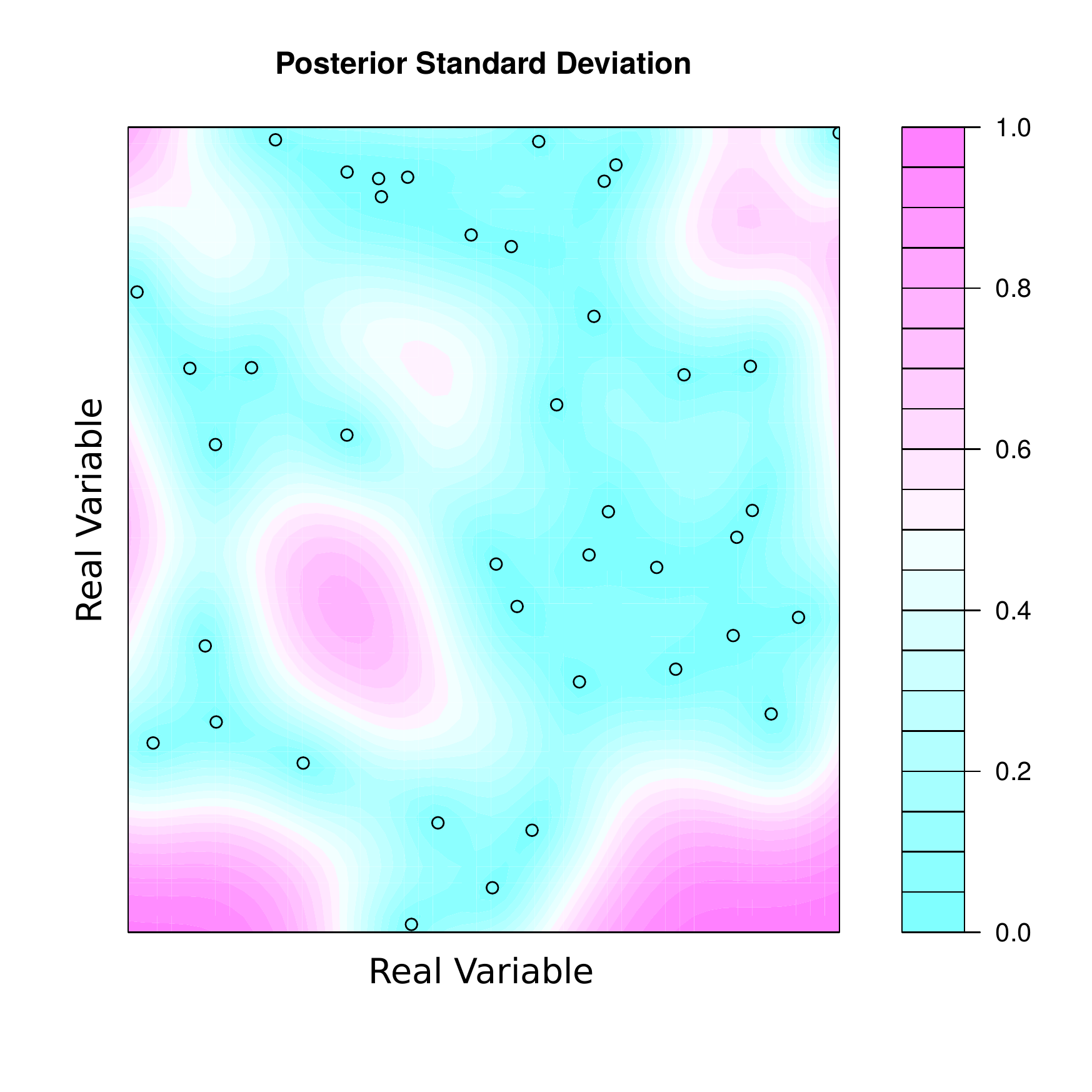} \\
\end{tabular}
\end{center}
\vspace{-1.4cm}
\caption{{\small (top) Posterior mean and standard deviation of a GP model over a 2-dimensional space in which the first dimension
can only take $5$ different integer values and when the covariance function in (\ref{eq:covariance}) is used. Note that the second 
dimension can take any real value. (bottom) Same results for a GP model using a covariance function without the proposed
transformation.  Best seen in color.}}
\label{fig:posterior}
\vspace{-0.75cm}
\end{figure}

Figure \ref{fig:posterior} illustrates the modelling properties of the proposed covariance function
(\ref{eq:covariance}). It shows the mean and standard deviation of the posterior distribution given some 
observations. It compares results with a standard GP that does not use the proposed transformation. In this case 
the data has been sampled from a GP using the covariance function in (\ref{eq:covariance}) with $k(\cdot,\cdot)$ 
the squared exponential covariance function \citep{rasmussen2005book}. One dimension takes continuous vales and 
the other values in $\{0,1,2,3,4\}$. Note that the posterior distribution captures the constant behavior 
of the function in any interval of values that are rounded to the same integer, only for the integer 
dimension (top). A standard GP (corresponding to the basic approach in Section \ref{sec:dealing}) cannot capture this (bottom).

\section{Experiments}

We compare the performance of the proposed approach for BO with the basic approach 
described in Section \ref{sec:dealing}. Each method has been implemented in the software for BO Spearmint. We use a Mat\'ern 
covariance function, and estimate the GP hyper-parameters using slice sampling \citep{murray2010}. The 
acquisition function employed is EI.

\begin{figure}[htb]
\begin{tabular}{cc}
        \includegraphics[width=0.475\linewidth]{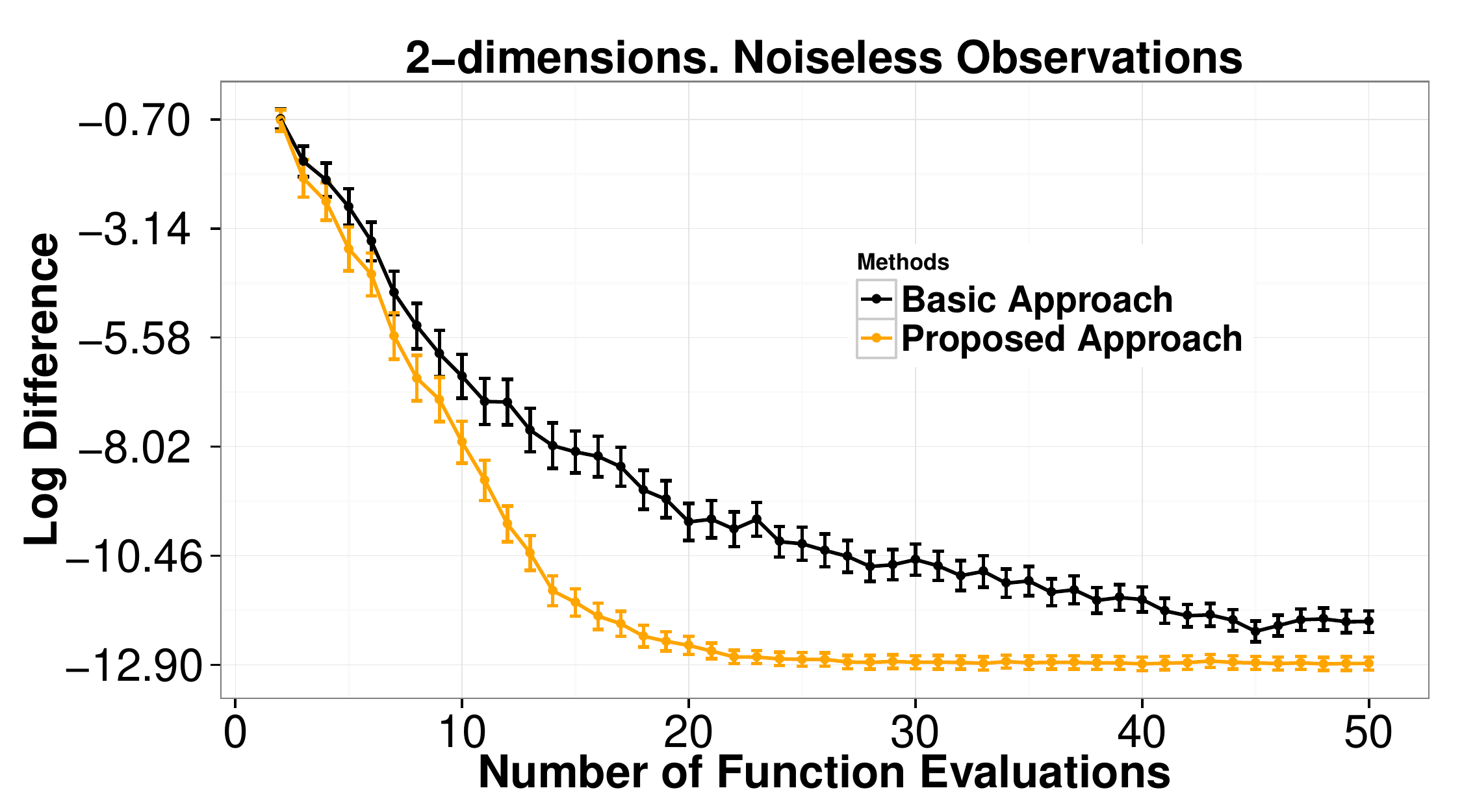} &
        \includegraphics[width=0.475\linewidth]{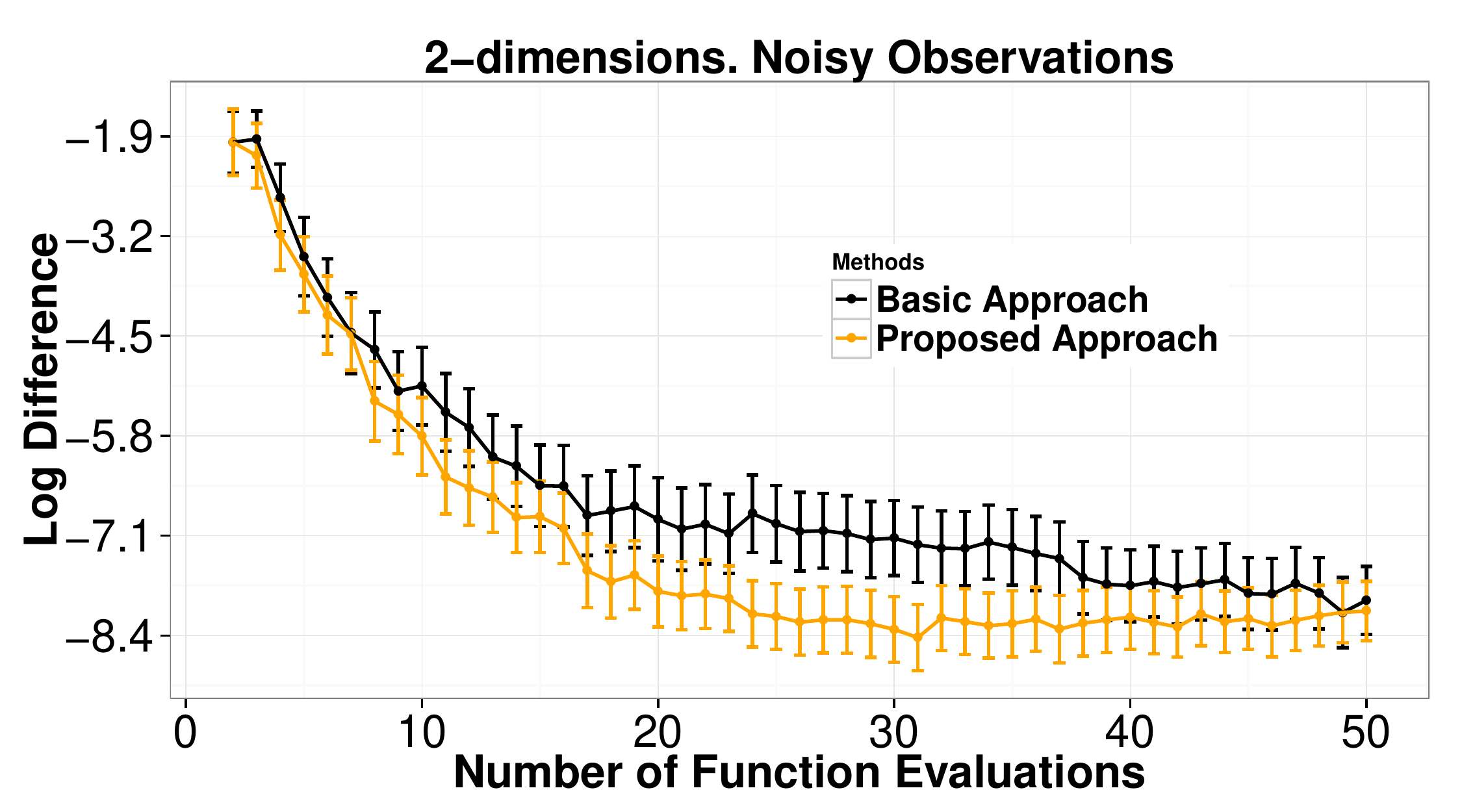} \\
        \includegraphics[width=0.475\linewidth]{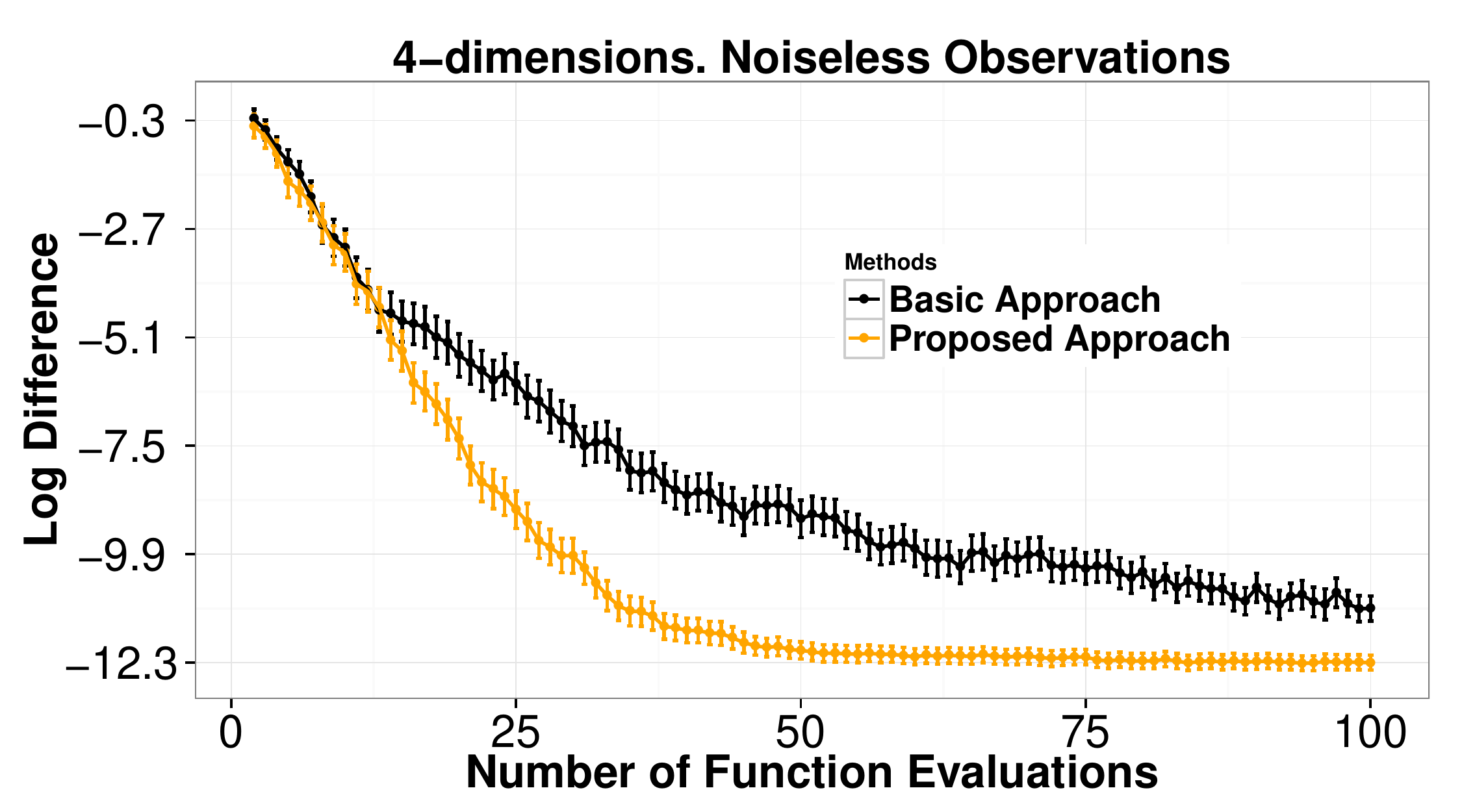} & 
        \includegraphics[width=0.475\linewidth]{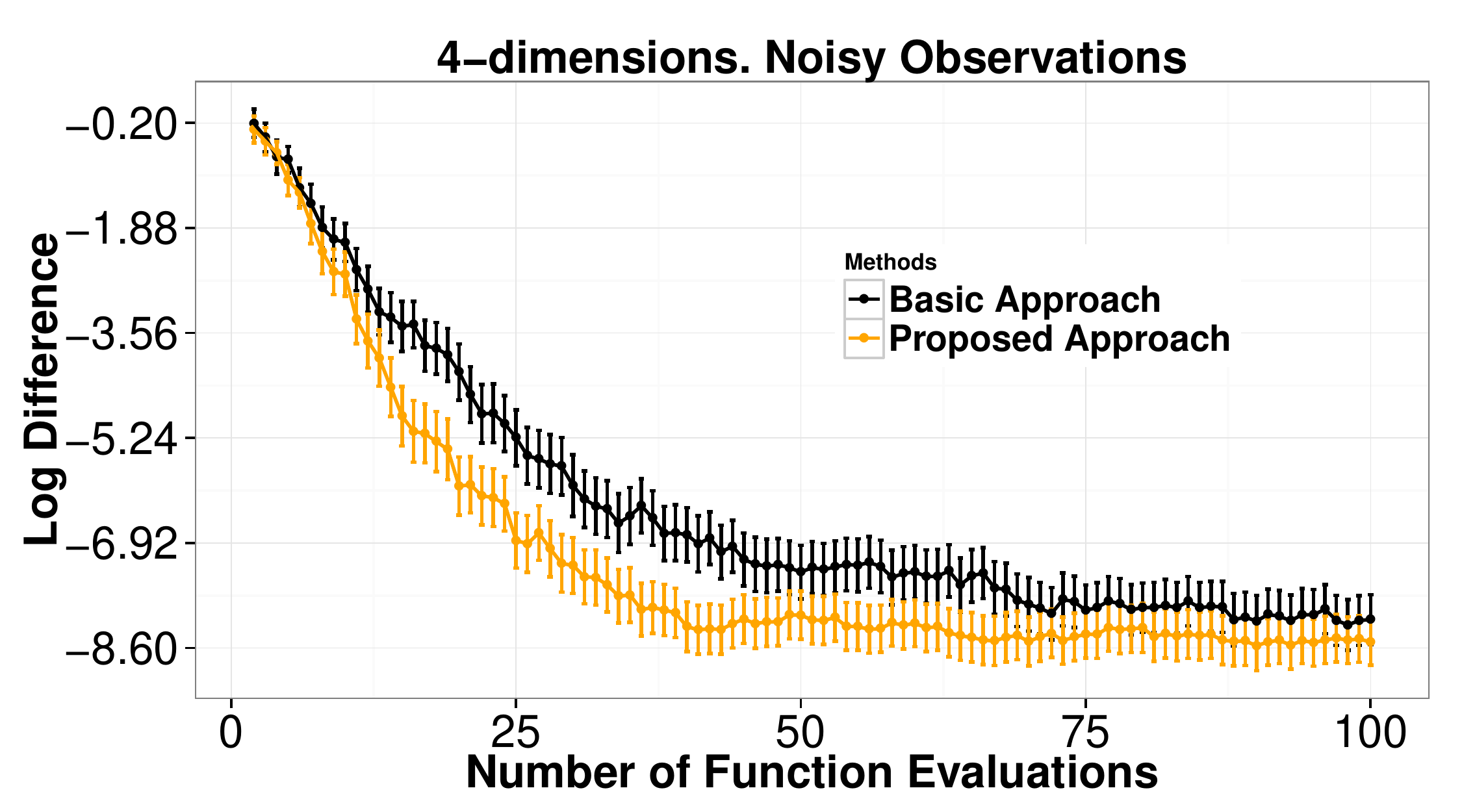} \\
\end{tabular}
\vspace{-.5cm}
\caption{{\small Average results on the synthetic experiments with 2 and 4 dimensions.}}
\label{fig:results_synthetic}
\vspace{-.15cm}
\end{figure}

A first batch of experiments considers two synthetic objectives. The first objective depends on 2 variables. In this case the first 
variable takes values in the interval $[0,1]$, and the second variable takes values in the set $\{0,1,2\}$. The second objective 
depends on 4 variables. In this case the first 2 variables take values in the interval $[0,1]$. The remaining 2 variables take 
values in the set $\{0,1,2,3\}$ and $\{0,1,2\}$, respectively. In each case, we sample the objectives from a GP prior 
using (\ref{eq:covariance}) as the covariance function. We run each BO method (proposed and basic) for 50 and 100 iterations,
in the case of each objective, and report the logarithm of the distance to the minimum value of each objective as a 
function of the evaluations done. We consider 100 repetitions of the experiments. We also consider these experiments when
the objectives are contaminated with additive Gaussian noise with variance $0.01$ and $0.001$, respectively.
The results obtained are displayed in Figure \ref{fig:results_synthetic}. We observe that the proposed approach gives
better results than the basic approach. In particular, it finds points that are closer to the optimal one with a smaller
number of evaluations of the objective, both in the case of noisy and noiseless evaluations. 

\begin{figure}[htb]
	\begin{center}
        \includegraphics[width=0.6\linewidth]{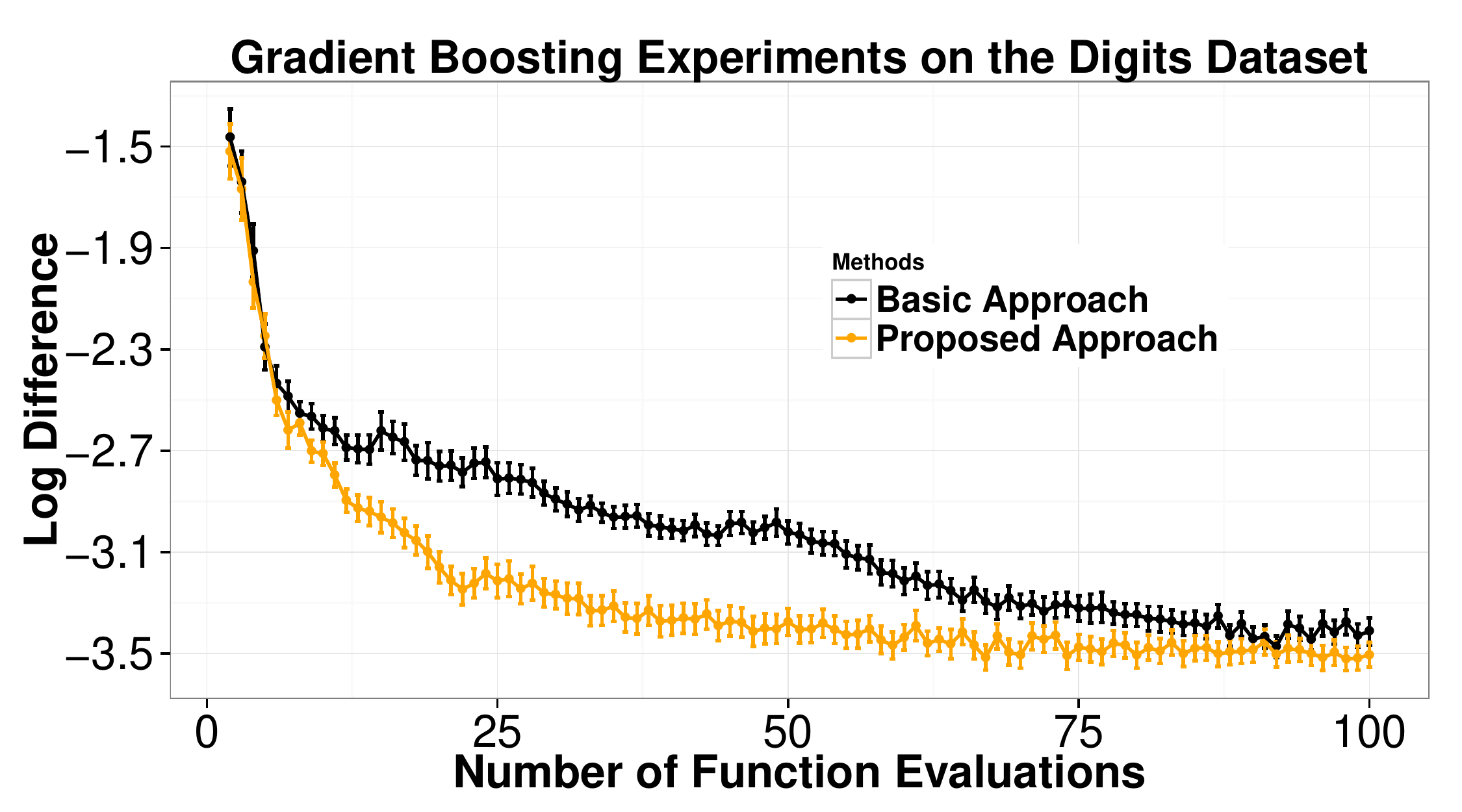} 
	\end{center}
\vspace{-.5cm}
\caption{{\small Average results on the digits dataset using gradient boosting.}}
\label{fig:results_digits}
\vspace{-.15cm}
\end{figure}

A last batch of experiments considers finding the learning rate and maximum tree depth
of a gradient boosting ensemble \citep{friedman2001greedy} of 100 trees that leads to the best 
performance on the digits dataset from the scikit-learn python package \citep{scikit-learn}. 
We use 70\% of the data for training and 30\% for validation and compute the performance in terms of 
the test log-likelihood. We consider optimizing the logarithm of the learning rate and five different
values for the maximum depth of the trees. Namely $\{1,2,3,4,5\}$. Again, we report the logarithm
of the distance to best observed performance on the validation set. We run each BO method 
(proposed and basic) for 100 iterations. We consider 100 repetitions of the experiments.
The results obtained are displayed in Figure \ref{fig:results_digits}. Again, the proposed approach gives better 
results than the basic approach. More precisely, it is able to find ensembles with better prediction properties 
on the validation set with a smaller number of evaluations.

\section{Conclusions}

We have described a new approach to deal with integer-valued variables in BO methods using GPs
as the underlying model. This approach consists in (i) rounding real values to integer values and (ii) 
modifying the covariance function of the GP to account for the fact that the objective should be constant
in the interval of all the real values that are rounded to the same integer value. The proposed
approach has been evaluated in both synthetic and real problems and compared with a basic approach
to handle integer-valued variables. These experiments show that the better modeling properties of 
the proposed approach lead to better results. More precisely, a BO method using such an approach 
finds solutions that are closer to the optimal ones in a smaller number of iterations.

\section*{Acknowledgments}

The authors gratefully acknowledge the use of the facilities of Centro
de Computaci\'on Cient\'ifica (CCC) at Universidad Aut\'onoma de
Madrid. The authors also acknowledge financial support from Spanish
Plan Nacional I+D+i, Grants TIN2013-42351-P, TIN2016-76406-P,
TIN2015-70308-REDT and TEC2016-81900-REDT (MINECO/FEDER EU), and from
Comunidad de Madrid, Grant S2013/ICE-2845.

\bibliography{references}

\end{document}